\documentclass[10pt,journal]{IEEEtran}
\usepackage{amsmath,amsfonts}
\usepackage{algorithmic}
\usepackage{algorithm}
\usepackage{array}
\usepackage[caption=false,font=normalsize,labelfont=sf,textfont=sf]{subfig}
\usepackage{textcomp}
\usepackage{stfloats}
\usepackage{verbatim}
\usepackage{graphicx}
\usepackage{cite}
\usepackage{url}
\usepackage{hyperref}
\usepackage{tabu}
\usepackage{threeparttablex}
\usepackage{xcolor}
\usepackage[subtle,tracking=normal]{savetrees}
\newcommand{\smalltt}[1]{{\small\texttt{#1}}}

\begin{document}
\bstctlcite{IEEEexample:BSTcontrol}
\title{From Vocal Instructions to Household Tasks: The Inria TIAGo++ in the euROBIN Service Robots Coopetition}

\author{
Fabio Amadio$^{*}$, Clemente Donoso$^{*}$, Dionis Totsila$^{*}$, Raphael Lorenzo, Quentin Rouxel, Olivier Rochel, Enrico Mingo Hoffman, Jean-Baptiste Mouret, Serena Ivaldi 
\thanks{
This work was supported by the EU Horizon project euROBIN (GA n.101070596), the France 2030 program through the PEPR O2R projects AS3 and PI3 (ANR-22-EXOD-007, ANR-22-EXOD-004), the Agence Innovation Defense (ATOR project), the CPER CyberEntreprises, and the Creativ'Lab platform of Inria/LORIA. \\ 
All the authors are with Inria, Universit\'{e} de Lorraine, CNRS, 54000 Nancy, France. \; ($^{*}$) Equal contribution.
}
}

\maketitle

\begin{abstract}
This paper describes the Inria team’s integrated robotics system used in the 1st euROBIN \textit{coopetition}, during which service robots performed voice-activated household tasks in a kitchen setting. The team developed a modified TIAGo++ platform that leverages a whole-body control stack for autonomous and teleoperated modes, and an LLM-based pipeline for instruction understanding and task planning. The key contributions (opens-sourced) are the integration of these components and the design of custom teleoperation devices, addressing practical challenges in the deployment of service robots.
\end{abstract}


\section{Introduction}
This paper describes the system integration and the software/hardware modules used by the Inria team participating in the 1st \href{https://www.eurobin-project.eu/index.php/competitions/coopetitions}{euROBIN \textit{coopetition}} (i.e., cooperative competition, where teams are rewarded when they collaborate in sharing software), which took place in Nancy, France, on the 25-28 November 2024.
\href{https://www.eurobin-project.eu/}{EuROBIN} is a Network of Excellence in AI and Robotics, funded by the European Commission. Among its objectives, it promotes transfer of robotics and AI software, methods and practices, by organizing annual robotics events where several teams collaborate to solve challenging tasks.
Twenty teams participated in the 1st \textit{coopetition}, in three different leagues.
The Inria team participated in the \textit{Service Robots League}, including six teams/robots, where mobile manipulators interact with people and objects in a domestic environment.
Service or domestic robots must possess navigation and manipulation skills, as well as the ability to understand and interpret commands from humans, and act accordingly~\cite{zhang2023llmhrireview}. Localization, perception and control are critical to navigate in a cluttered environment and interact with objects: doing this robustly in environments different from the lab is still challenging. For interaction, Large Language Models (LLMs) recently showed great potential in connecting natural language to robotic actions~\cite{song2023llm,ren2023robotsaskhelpuncertainty}, relying on ``common sense'' reasoning to comprehend ambiguous instructions; but they must be both reliable and fast enough for real-time human-robot interaction.

To help teams address these problems, the euROBIN \textit{coopetition} introduced a simplified kitchen scenario in which the robot was requested to understand and execute standard instructions following two patterns: (1) \textit{pick an object from a designated location and place it at another location}; (2) \textit{pick an object from a designated location and deliver it to a person}.
Teleoperation was allowed (but penalized in the point system) to extend the use of the platforms to new or unexpected situations and cope with failures that might occur during autonomous operation. 
The \emph{coopetition} participants developed both hardware and software, addressing the challenges related to the system integration, including third-party software integration, and execution in realistic competition setting.

\begin{figure}[t]
    \centering
    \includegraphics[width=0.9\linewidth]{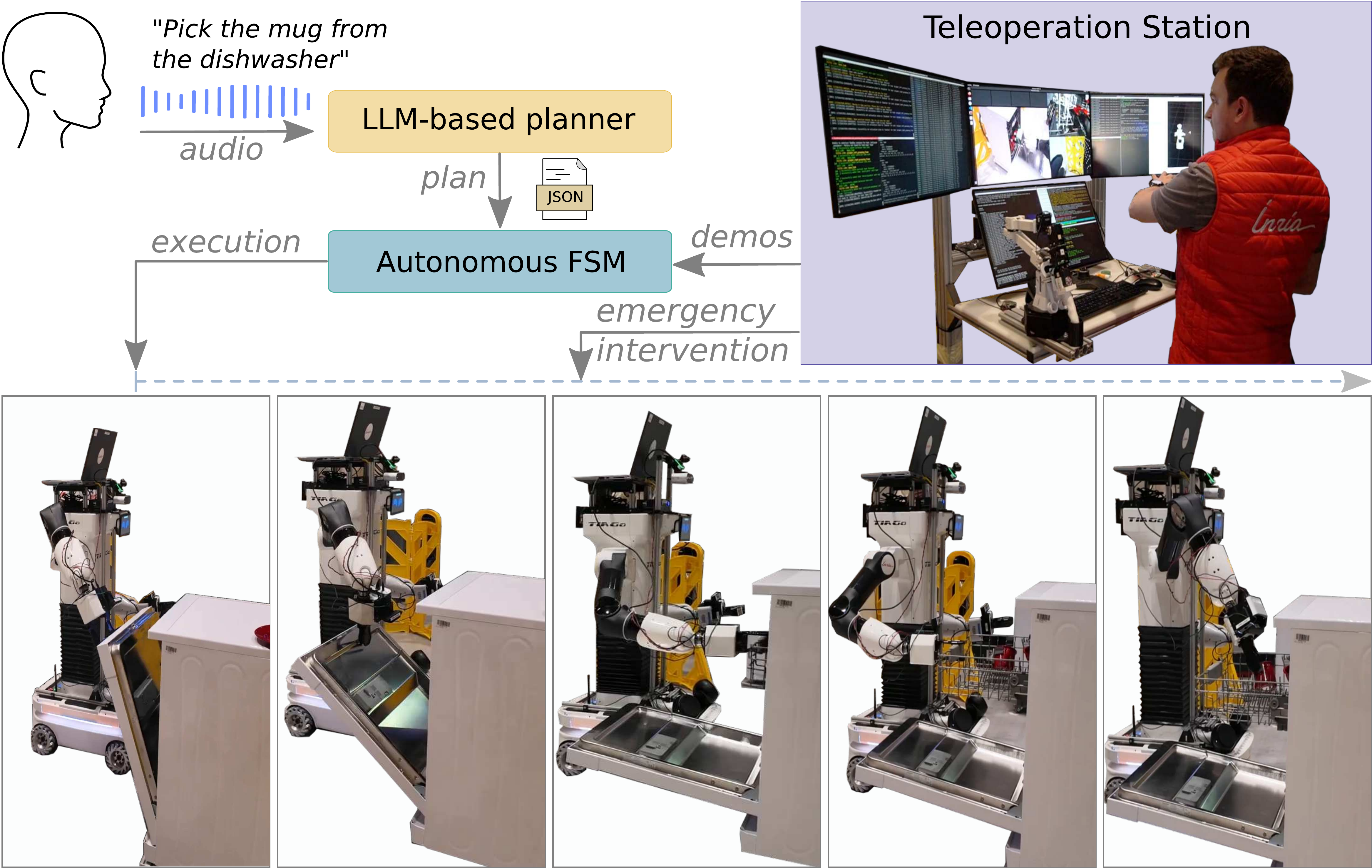}
    \caption{System overview: our LLM-based planner understands voice commands and generate a plan (in JSON format) that is used to assemble a Finite State Machine (FSM) for carrying out the instruction. Teleoperation is used both to record expert demonstrations, and to intervene in case of emergency or failure.}
    \label{fig:concept}
\end{figure}

\begin{figure*}[!bhtp]
    \centering
    \includegraphics[width=0.85\textwidth]{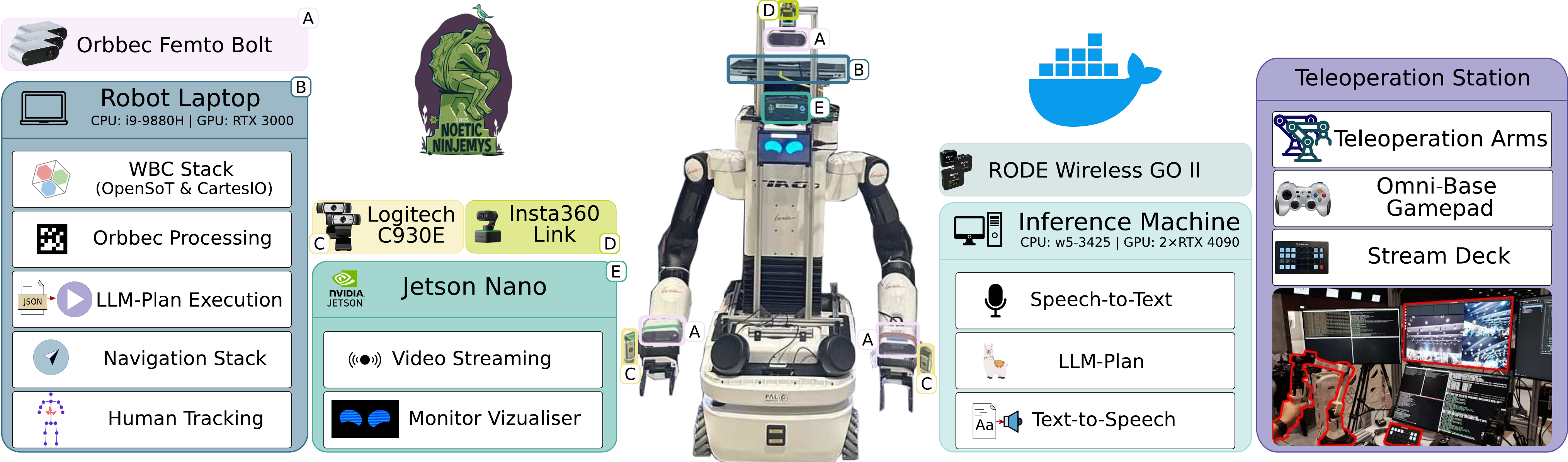}
    \caption{General overview of our system, based on a dual-arm TIAGo++ robot with an omnidirectional mobile base. Each block includes its connected peripherals, positioned above it. The robot is equipped with three RGB-D cameras (one fixed and two mounted on the grippers) and three webcams for teleoperation. A laptop mounted on the robot manages the WBC stack (Sec.~\ref{subsec:cartesio}), RGB-D camera processing (Sec.~\ref{subsec:tags-detector}), plan execution (Sec.~\ref{subsec:deployment}), navigation (Sec.~\ref{subsec:navigation}) and human pose tracking (Sec.~\ref{subsec:human-tracking}). Additionally, a Jetson Nano on the robot streams webcam footage to the teleoperation station and renders interactive visuals on a 7-inch screen mounted on the robot. Computationally intensive tasks, such as LLM-based plan generation (Sec.~\ref{subsec:speech}), are offloaded to a remote machine equipped with two GPUs. Finally, the Teleoperation Station (Sec.~\ref{subsec:teleop}) serves as a platform for (a) collecting demonstrations and (b) controlling the robot during failures or emergencies. We employed ROS Noetic middleware to integrate the different units (all running in dedicated Docker containers), connecting them to the TIAGo++'s internal PC (where the \smalltt{roscore} is running).}
    \label{fig:system-overview}
    \vspace{-0.5cm}
\end{figure*}

The Inria's robot (Fig.~\ref{fig:concept}) is a modified \href{https://pal-robotics.com/robot/tiago}{TIAGo++} with omnidirectional base. Its main components are a Whole-Body Control (WBC) stack for both teleoperated and autonomous operation and a LLM-based plan generation pipeline for instruction understanding. 
Together with the description of the system components and their integration (Fig.~\ref{fig:system-overview}), we share the software and the design of the bimanual teleoperation devices (links reported in Table \ref{table:links}).
A video of the robot in action is available at \href{https://youtu.be/5mSIYuH4Mdk}{\smalltt{youtu.be/5mSIYuH4Mdk}}.

\section{Modules and Components}
\subsection{Motion control}\label{subsec:cartesio}
This module coordinately controls all the TIAGo++ DOFs through an optimization-based WBC~\cite{wbc:ram19} stack that reads the proprioceptive data from the robot and maps user-level Cartesian references into joint-space commands for the low-level controller. This WBC formulation offers a principled and unified solution to handle the redundancy of the platform (e.g., combining arm motions and torso motions), while taking into account joint position and velocity limits and avoiding self-collisions.
This WBC stack is based on the CartesI/O~\cite{cartesio:icra19} framework and the OpenSoT~\cite{opensot:ram24} library, which enables instantaneous control by formulating and solving Quadratic Programming (QP) problems using atomic entities like \emph{Tasks} and \emph{Constraints}.  
CartesI/O automates the setup of OpenSoT problems via configuration files and provides interfaces for interacting with tasks and constraints, offering APIs in C++ and Python, and supporting frameworks like ROS.

The control problem consists of three Cartesian tasks, formulated at the velocity level and arranged in a soft priority hierarchy: the left and right arm tasks, and the control of the omni-directional base. In the null-space, a postural task is employed to stabilize the robot's self-motions around the desired nominal configuration. 
The control of the base is achieved by modeling the robot as a floating-base system, constraining the omni-directional motion only on the ground. 
\par
We employ an open loop control scheme initialized at the initial references in generalized coordinates $\left[\mathbf{Q}_{r,0}^T, \ \mathbf{q}_{r,0}^T\right] \in SE(3) + \mathbb{R}^{19}$ retrieved when the controller is started. At every control loop, the model is updated with the integrated output of the QP, namely $\left[\boldsymbol{\nu}^T \ \dot{\mathbf{q}}_d^T\right] \in \mathbb{R}^{6+19}$, as well as the postural task at the secondary priority level. 
The Cartesian reference commands are defined using homogeneous poses, $\mathbf{T}_r \in \mathbb{R}^{4 \times 4}$. A properly computed orientation error, specifically the quaternion error, is calculated with respect to the forward kinematics. This error is then multiplied by a positive definite gain matrix, $\mathbf{K}_C \in \mathbb{R}^{6 \times 6}$, and combined with feed-forward Cartesian velocities, $\mathbf{v}_r \in \mathbb{R}^3$ for linear motion and $\boldsymbol{w}_r \in \mathbb{R}^3$ for angular motion. Similar considerations are done for the lower priority postural task, where the desired posture $\mathbf{q}_d \in \mathbb{R}^{15}$ can be adjusted during task execution. 
Collision avoidance is not explicitly handled in the current implementation; however, it can be naturally incorporated into the WBC formulation, as shown in \cite{totsila2025safe}.
\par
The WBC scheme is implemented as a ROS node running at $250 \ \text{Hz}$ using the CartesI/O API, with the underlying optimal control problem solved in approximately 1–2 ms. The output of this node is retrieved by another node, namely \smalltt{ros\_control\_bridge}, in charge of dispatching the solution, meaning  sending joint positions commands through a ROS topic interface as a \smalltt{JointTrajectory} message to each kinematic chain controller, which is part of the \smalltt{ros\_control} layer of TIAGo++. Meanwhile, base velocities are sent as \smalltt{Twist} messages to the \smalltt{cmd\_vel} topic of the base controller.
Separating upper-body joint references from those of the wheels may introduce challenges in real-time execution and in maintaining synchronization between upper-body and base motions. Nevertheless, this approach proved adequate for tasks executed at moderate velocities.

\subsection{Teleoperation}
\label{subsec:teleop}
The teleoperation interface is used both to record demonstrations (Sec.~\ref{subsec:teaching}) and as a fallback mode when the autonomous mode fails.
It is designed around two low-cost master arms (Dynamixel actuators, used passively as encoders) inspired by the Aloha project \cite{aloha:fu2024mobile}, whose end-effector position (forward kinematics) is computed with the Pinocchio library \cite{8700380}. Each of them consists of seven motors: six control the full pose of the end-effector, while the seventh commands the opening and closing of the gripper. This configuration effectively maps the six DOFs required to perform all necessary motions despite the DOFs asymmetry~\cite{9131861}. However, it introduces mismatches between joint limits of the teleoperation device and the robot and can reduce the workspace compared to that of the robot.

A gamepad is used to send angular and linear velocity commands to the mobile base.  All the commands are sent to CartesI/O through its Python ROS client at  $100 \ \text{Hz}$. 

Three cameras are used: one mounted on each end-effector (\href{https://www.logitech.com/fr-fr/products/webcams/c930e-business-webcam.960-000972.html?sp=1&searchclick=logi}{Logitech C930E} and a 3-DOF controllable orientation camera \href{https://www.insta360.com/product/insta360-link}{Insta 360 Link}, positioned as the ``head''. The video is streamed over UDP using a \href{https://gstreamer.freedesktop.org/}{GStreamer} pipeline that leverages the on-camera hardware H.264 encoder, resulting in an end-to-end latency of approximately $30--40 \ \text{ms}$. Finally, a \href{https://www.elgato.com/us/en/p/stream-deck-xl}{Stream Deck XL} is used to activate standard routines (e.g., homing) with the press of hardware buttons.   \\
\textbf{Practical considerations: }
The system is designed to run on minimal hardware with limited software dependencies; a Dockerfile and installation instructions are provided in the repository. During assembly, all motors must be set to zero before mounting the arm to ensure proper calibration (see supplementary assembly guide). The teleoperation mapping tracks the master arm orientation, while its position is computed through an offset, making it independent of the initial control position. Both wired and wireless setups are supported. Wi-Fi enables untethered operation but requires a non-congested frequency band to maintain reliable teleoperation; in practice, network saturation can increase latency beyond $200 \ \text{ms}$, therefore an Ethernet connection is recommended in such settings.

\subsection{Object pose estimation}
\label{subsec:tags-detector}
The robot needs to estimate the 6D poses of objects and key elements in the environment from RGB-D images for manipulation and navigation. We relied on AprilTag\cite{wang2016apriltag} fiducial markers. Although this approach is a substantial simplification of the problem, it is a reliable pose estimation module to integrate in our system, that will be eventually easily replaced by a 6D-pose estimation module in the future (e.g. ~\cite{labbe2022megapose}).

Our system relies on two RGB-D cameras (\href{https://www.orbbec.com/products/tof-camera/femto-bolt/}{Orbecc Femto Bolt}), different from the cameras used for teleoperation: one on the left wrist (for low-range detection), and one on top of the torso (for long-range detection).
The tag IDs and the pixel coordinates of the four corners are extracted from the camera’s color images using the AprilTag library~\cite{krogius2019flexible}. Contrary to classic tag-based tracking, the tags are identified on the RGB image, but the 3D position and orientation of each marker in the camera frame are computed using the ordered point cloud from the camera's depth (time-of-flight) sensor. Each marker pose is then calculated and broadcasted on the ROS \smalltt{tf} tree, which makes it possible to combine it with the forward kinematics of the robot to express the position in the base frame. In our experiments, this RGB-D approach provided more accurate and less noisy estimations, particularly for orientation and distance, than relying solely on AprilTag processing of color images. 

\subsection{Teaching object-centric manipulation skills}\label{subsec:teaching}
Many of the considered manipulation tasks (e.g., opening the dishwasher or the cabinet) require complex end-effector trajectories that are difficult to program explicitly but can be effectively reproduced via skill demonstrations. During teleoperation, we record the end-effector trajectory, gripper commands, the target object's AprilTag pose, and the relative pose between the robot base and the object. In post-processing, the demonstration is expressed in the object reference frame, similarly to~\cite{amadio2022target}.
During autonomous execution, the object pose is estimated, the robot base is aligned with the demonstrated base–object offset, and the trajectory is transformed into the base frame based on the detected pose before being replayed. This removes assumptions about object placement and allows the task to be reproduced under moderate pose variations. The gripper command is replayed in synchronization with the end-effector motion, ensuring consistent grasp execution.
This approach proved sufficiently robust while relying on a single high-quality demonstration per task. An image-based policy learning system~\cite{aloha:fu2024mobile} could replace this pipeline, potentially generalizing to more diverse scenarios, at the cost of requiring significantly more demonstrations.\\
\textbf{Practical considerations}: Proprioceptive data, gripper status, and AprilTag pose are recorded at $\sim100 \ \text{Hz}$. Post-processing includes start/end selection, expressing the trajectory in the AprilTag frame at the initial time, interpolation at $50 \ \text{Hz}$, and optional temporal scaling. During replay, the end-effector is first positioned at the starting pose in the current tag frame.

\subsection{Human tracking}\label{subsec:human-tracking}
To autonomously carry out the handover instruction, we detect and track the person in the kitchen from the RGB-D camera stream (torso camera).
Our approach involves five steps: (1) detecting humans, (2) tracking them, (3) estimating their 2D position in the RGB image, (4) computing the 3D coordinates using depth, and (5) determining if the persons are attentive (i.e., are looking at the robot) by checking their head pose. The closest attentive individual is selected as the handover target and their pose is broadcasted on the ROS \smalltt{tf} tree.

Human detection and pose estimation were performed using YOLOv3~\cite{redmon2018yolov3} for bounding box detection and ViTPose~\cite{xu2022vitpose} for human pose estimation, following the standard MMDetection~\cite{mmdetection} and MMPose~\cite{mmpose2020} pipeline. For tracking, we utilized the IoU-based tracker from MMPose with a threshold of $0.3$, running at $10$ fps. We filtered detections based on bounding box size to mitigate latency issues in pose estimation.

We used 6DRepNet ~\cite{9897219} to estimate the head's 6D pose as a proxy for gaze direction, given its robustness under challenging conditions compared to gaze estimation methods that depend on high-resolution eye images and favorable lighting conditions.
Since the model is designed for front-facing heads, we applied it only when a rough heuristic indicated that the person was facing the camera. This was determined by the pixel difference between the detected left-ear and right-ear keypoints (greater than a threshold of $30$ pixels).

\subsection{Navigation}\label{subsec:navigation}
We relied on a basic navigation node that implements a ROS \smalltt{MoveBaseAction} action to move the base at the desired offset w.r.t. a reference frame (e.g., AprilTag, tracked human) present in the \smalltt{tf} tree.
Together with simple, ad-hoc obstacle detection using the onboard LiDARs and target searching procedures (the robot rotates in place until the desired AprilTag is detected), this solution proved sufficient in the context of the \textit{coopetition} and eliminates the need to build a map of the environment. In future developments, it would be possible to replace this simple solution with state-of-the-art navigation algorithms~\cite{macenski2023survey} without modifying the overall system architecture.

\subsection{LLM-based plan generation}
\label{subsec:speech}
In the euROBIN \textit{coopetition}, at the beginning of each run, a referee communicates a randomly generated task to the robot. The robot must interpret the verbal instruction and formulate a corresponding task execution plan.
While structured human commands can be handled using traditional approaches like SnipsNLU~\cite{coucke2018snips}, the use of LLMs offers a significant advantage in processing more general, unstructured inputs expressed by a human in natural language. LLMs can infer well-structured outputs even from ambiguous commands, deal with synonyms or paraphrases, and their capability to generate extra textual fields, such as chain-of-thought reasoning~\cite{wei2023chainofthought} improves interpretability.

We integrate Faster-Whisper~\cite{raford2023whisper} for speech-to-text, passing transcribed commands to a Llama-3.1-8B-Instruct model~\cite{grattafiori2024llama3herdmodels} via a task-specific system prompt and few-shot examples~\cite{brown2020language}. 
Rather than reactive tool-calling, we generate a complete structured JSON plan that allows the system to validate the action sequence and preconditions before execution. The inclusion of a \textit{chain-of-thought} field serves a dual purpose: ensuring the plan adheres to task constraints through step-by-step reasoning~\cite{wei2023chainofthought}, and providing a reasoning trace that is verbally communicated to the user via Coqui TTS~\cite{Eren_Coqui_TTS_2021} to enhance HRI transparency.

\begin{figure}[t]
    \centering
    \includegraphics[width=0.90\linewidth]{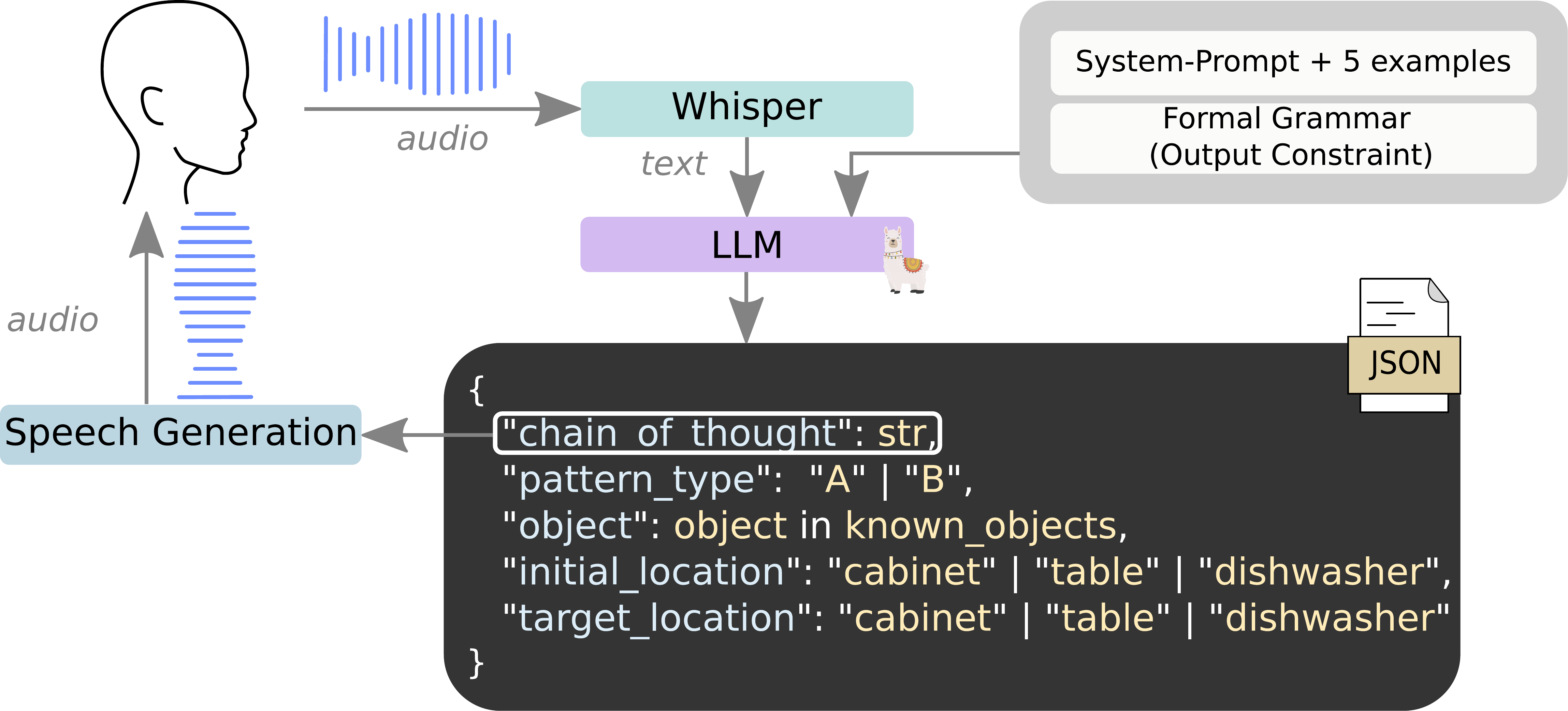}
    \caption{LLM-based plan generation overview (cf. Sec.~\ref{subsec:speech}).}
    \label{fig:llm-json}
\end{figure}

Additionally, we use \smalltt{llama-cpp} to constrain the LLM output through grammar-based token sampling~\cite{willard2023efficient}. Unlike standard zero-shot tool usage, this strictly ensures 100\% adherence to the execution schema, preventing hallucinated arguments or syntax errors common in local, quantized deployments. This methodology ensures a consistently parsable JSON output describing the plan (Fig.~\ref{fig:llm-json}).

\textbf{Practical considerations on system requirements: }
Efficient inference requires access to high-performance GPUs. In our setup, running Faster-Whisper, Llama-3.1-8B-Instruct, and ljspeech/vits requires approximately $16.1 \ \text{GB}$ of VRAM. Using older GPUs or CPU inference leads to significantly slower response times. Over $50$ trials, LLM inference on our GPU had a median latency of $2.51$ s (25th–75th percentile: $2.40$–$2.65$ s).
\section{Integration and Deployment}
\label{subsec:deployment}
All the described components were deployed using Docker containers in the robot laptop (for CartesI/O, AprilTag detection, human tracking), the teleoperation station (for devices management and communication), the Jetson Nano (for streaming webcams), and the inference machine (for microphone management, speech-to-text, LLM-plan, and text-to-speech networks). They communicate using ROS (Noetic).

We relied on the \smalltt{smach} library to configure and execute robot behaviours as Finite-State Machines (FSMs). In particular, the CartesI/O Python ROS client sends commands to the WBC, and the standard ROS actions or services to send goals to the navigation node, or control the grippers.
We employed two basic types of \smalltt{smach} states for motion: \emph{way-points} (where the indicated end-effector follows a sequence of target points) or \emph{demo-playback} (where the end-effector replicate a demonstrated trajectory recorded via teleoperation, Sec.~\ref{subsec:teaching}).
In both cases, the user is able to specify the reference frame in which the commanded trajectory is expressed, making it immediate to define motions as offsets from a particular frame of interest (e.g., AprilTag or tracked human).
Such states are the main building blocks of the a set of sub-FSMs that solve individual portions of the instruction and that can be assembled together. Specifically, these are: \smalltt{pick\_from\_<loc>(object)}, \smalltt{place\_at\_<loc>(object)}, \smalltt{go\_to(loc)}, and \smalltt{handover} (where \smalltt{loc} and \smalltt{object} belong to predefined sets). 
Different motion specifications (targets, times, demos, etc.) are stored in a configuration file and labelled following a uniform pattern.

Hence, we employ an ``orchestrator'' node (running on the CartesI/O container) that requests the plan in JSON format from the LLM (via a ROS service call), and uses it to generate programmatically (by concatenating sub-FSMs) the correct FSM and run it to carry out the instruction. 
Throughout the \textit{coopetition}, in case the robot was unable to successfully complete the assigned task, teleoperation served as a fallback solution to handle failure cases, with intervention decisions made by the supervising operator.

\section{Discussion}
Our LLM-based plan generation pipeline robustly understood diverse voice commands from different users with varying accents, consistently structuring plans correctly despite phrasing variations. In the competition trials, command interpretation achieved 7/7 successes, confirming the reliability of the language-to-plan interface in the tested scenarios.
AprilTag-based navigation achieved 17/18 successes in the structured setup but required extensive environmental instrumentation, limiting scalability. In contrast, marker-based object pose estimation was less robust (1/3 successes) and struggled with arbitrary object placement and cluttered scenes, which constrained fully autonomous manipulation. Overall, manipulation achieved 12/15 successes (10/13 teleoperated, 2/2 autonomous). While teleoperation enabled diverse tasks, it required significant expertise; improving workspace and kinematic constraints would enhance usability. We also plan to implement the teleoperation architecture in C++ to further enhance performance.
\textit{Demo-playback} proved effective for performing complex interactions, such as opening a dishwasher. However, it lacks generalization capabilities, for example, to different dishwasher types. The existing demonstration recording pipeline can be leveraged to collect datasets for training more robust and generic policies based on diffusion~\cite{chi2023diffusion} or flow matching~\cite{rouxel2024flow}, which could handle a wider variety of environments.
Upgrading the perception and navigation modules with state-of-the-art object tracking approaches \cite{labbe2022megapose} therefore represents an essential direction for future work.
LLM-based planning, currently limited to instruction processing, could be extended to incorporate execution-time feedback for adaptive re-planning.

\begin{table*}[ht]
\begin{ThreePartTable}
\centering
\begin{tabu} to \linewidth {|X[1.3,l]|X[3,l]|}
 \hline
 \textit{Software \& Hardware Modules} & \textit{Links} \\
 \hline
  OpenSoT                &  \href{https://github.com/ADVRHumanoids/OpenSoT}{\texttt{https://github.com/ADVRHumanoids/OpenSoT}} \\ 
  CartesI/O              &  \href{https://github.com/ADVRHumanoids/CartesianInterface}{\texttt{https://github.com/ADVRHumanoids/CartesianInterface}} \\ 
  TIAGo Dual CartesI/O configuration  &  \href{https://github.com/hucebot/tiago_dual_cartesio_config/tree/euRobin_nov24}{\texttt{https://github.com/hucebot/tiago\_dual\_cartesio\_config/tree/euRobin\_nov24}} \\
 \hline
 Teleoperation interface (with CAD files)        &  \href{https://github.com/hucebot/dxl_6d_input/tree/bimanual-teleoperation}{\texttt{https://github.com/hucebot/dxl\_6d\_input/tree/bimanual-teleoperation}} \\  
 StreamDeck controller        & \href{https://github.com/hucebot/stream_deck_controller}{\texttt{https://github.com/hucebot/stream\_deck\_controller}} \\ 
 \hline
 AprilTags detector      &  \href{https://github.com/hucebot/orbbec_apriltag_ros}{\texttt{https://github.com/hucebot/orbbec\_apriltag\_ros}} \\
 AprilTags generator     &  \href{https://github.com/hucebot/april_tag_generator}{\texttt{https://github.com/hucebot/april\_tag\_generator}}  \\
 \hline
 Human Tracking          & \href{https://github.com/hucebot/eurobin_human_tracking}{\texttt{https://github.com/hucebot/eurobin\_human\_tracking}} \\
 \hline
 LLM-based planner       & \href{https://github.com/hucebot/eurobin_llm_plan}{\texttt{https://github.com/hucebot/eurobin\_llm\_plan}} \\
\textit{Faster-Whisper}$^*$   & \href{https://github.com/SYSTRAN/faster-whisper}{\texttt{https://github.com/SYSTRAN/faster-whisper}} \\
 \textit{llama-cpp}$^*$         & \href{https://github.com/ggerganov/llama.cpp}{\texttt{https://github.com/ggerganov/llama.cpp}} \\
 \hline
 Finite State Machine for CartesI/O     & \href{https://github.com/hucebot/fsm_cartesio}{\texttt{https://github.com/hucebot/fsm\_cartesio}} \\
 \textit{smach}$^*$             & \href{https://wiki.ros.org/smach}{\texttt{https://wiki.ros.org/smach}}\\
 \hline
\end{tabu}
\begin{tablenotes}
    \item[*] Third-party software
\end{tablenotes}
\caption{Links to the used software and hardware components.}
\label{table:links}
\end{ThreePartTable}
\vspace{-0.5cm}
\end{table*}

\bibliographystyle{IEEEtran}
\bibliography{main}

@IEEEtranBSTCTL{IEEEexample:BSTcontrol,
  CTLuse_forced_etal       = "yes",
  CTLmax_names_forced_etal = "30",
  CTLnames_show_etal       = "2" 
}

@ARTICLE{wbc:ram19,
  author={Hoffman, Enrico Mingo and Caron, Stephane and Ferro, Francesco and Sentis, Luis and Tsagarakis, Nikos G.},
  journal={IEEE Robotics \& Automation Magazine}, 
  title={Developing Humanoid Robots for Applications in Real-World Scenarios [From the Guest Editors]}, 
  year={2019},
  volume={26},
  number={4},
  pages={17-19}
  }

@ARTICLE{opensot:ram24,
  author={Hoffman, Enrico Mingo and Laurenzi, Arturo and Tsagarakis, Nikos G.},
  journal={IEEE Robotics \& Automation Magazine (RAM)}, 
  title={The Open Stack of Tasks Library: OpenSoT: A Software Dedicated to Hierarchical Whole-Body Control of Robots Subject to Constraints}, 
  year={2024},
  volume={},
  number={},
  pages={2-12}
}

@INPROCEEDINGS{cartesio:icra19,
  author={Laurenzi, Arturo and Hoffman, Enrico Mingo and Muratore, Luca and Tsagarakis, Nikos G.},
  booktitle={IEEE International Conference on Robotics and Automation (ICRA)}, 
  title={CartesI/O: A ROS Based Real-Time Capable Cartesian Control Framework}, 
  year={2019},
  volume={},
  number={},
  pages={591-596}
 }

@inproceedings{aloha:fu2024mobile,
  author    = {Fu, Zipeng and Zhao, Tony Z. and Finn, Chelsea},
  title     = {Mobile ALOHA: Learning Bimanual Mobile Manipulation with Low-Cost Whole-Body Teleoperation},
  booktitle = {{Conference on Robot Learning (CoRL)}},
  year      = {2024},
}

@inproceedings{song2023llm,
  title     = {Llm-planner: Few-shot grounded planning for embodied agents with large language models},
  author    = {Song, Chan Hee and Wu, Jiaman and Washington, Clayton and Sadler, Brian M and Chao, Wei-Lun and Su, Yu},
  booktitle = {International Conference on Computer Vision (ICCV)},
  year      = {2023}
}

@inproceedings{ren2023robotsaskhelpuncertainty,
  title         = {Robots That Ask For Help: Uncertainty Alignment for Large Language Model Planners},
  author        = {Allen Z. Ren and Anushri Dixit and Alexandra Bodrova and Sumeet Singh and Stephen Tu and Noah Brown and Peng Xu and Leila Takayama and Fei Xia and Jake Varley and Zhenjia Xu and Dorsa Sadigh and Andy Zeng and Anirudha Majumdar},
  year          = {2023},
  booktitle = {Conference on Robot Learning (CoRL)}
  }

@article{zhang2023llmhrireview,
title = {Large language models for human–robot interaction: A review},
journal = {Biomimetic Intelligence and Robotics},
volume = {3},
number = {4},
pages = {100131},
year = {2023},
issn = {2667-3797},
doi = {https://doi.org/10.1016/j.birob.2023.100131},
author = {Ceng Zhang and Junxin Chen and Jiatong Li and Yanhong Peng and Zebing Mao},
}

@inproceedings{raford2023whisper,
author = {Radford, Alec and Kim, Jong Wook and Xu, Tao and Brockman, Greg and McLeavey, Christine and Sutskever, Ilya},
title = {Robust speech recognition via large-scale weak supervision},
year = {2023},
booktitle = {Proceedings of the 40th International Conference on Machine Learning (ICML)},
articleno = {1182},
}

@article{macenski2023survey,
  title={From the desks of ROS maintainers: A survey of modern \& capable mobile robotics algorithms in the robot operating system 2},
  author={Macenski, Steve and Moore, Tom and Lu, David V and Merzlyakov, Alexey and Ferguson, Michael},
  journal={Robotics and Autonomous Systems},
  volume={168},
  pages={104493},
  year={2023},
  publisher={Elsevier}
}

@inproceedings{
  xu2022vitpose,
  title={Vi{TP}ose: Simple Vision Transformer Baselines for Human Pose Estimation},
  author={Yufei Xu and Jing Zhang and Qiming Zhang and Dacheng Tao},
  booktitle={Advances in Neural Information Processing Systems},
  year={2022},
}

@inproceedings{redmon2018yolov3,
  title={Yolov3: An incremental improvement},
  author={Farhadi, Ali and Redmon, Joseph},
  booktitle={Computer Vision and Pattern Recognition ({CVPR})},
  volume={1804},
  pages={1--6},
  year={2018},
}

@misc{mmpose2020,
    title={OpenMMLab Pose Estimation Toolbox and Benchmark},
    author={MMPose Contributors},
    howpublished = {\url{https://github.com/open-mmlab/mmpose}},
    year={2020}
}

@article{mmdetection,
  title   = {{MMDetection}: Open MMLab Detection Toolbox and Benchmark},
  author  = {Chen, Kai and Wang, Jiaqi and Pang, Jiangmiao and Cao, Yuhang and
             Xiong, Yu and Li, Xiaoxiao and Sun, Shuyang and Feng, Wansen and
             Liu, Ziwei and Xu, Jiarui and Zhang, Zheng and Cheng, Dazhi and
             Zhu, Chenchen and Cheng, Tianheng and Zhao, Qijie and Li, Buyu and
             Lu, Xin and Zhu, Rui and Wu, Yue and Dai, Jifeng and Wang, Jingdong
             and Shi, Jianping and Ouyang, Wanli and Loy, Chen Change and Lin, Dahua},
  journal= {arXiv preprint arXiv:1906.07155},
  year={2019}
}

@INPROCEEDINGS{9897219,
  author={Hempel, Thorsten and Abdelrahman, Ahmed A. and Al-Hamadi, Ayoub},
  booktitle={2022 IEEE International Conference on Image Processing (ICIP)}, 
  title={6d Rotation Representation For Unconstrained Head Pose Estimation}, 
  year={2022},
  volume={},
  number={},
  pages={2496-2500},
  doi={10.1109/ICIP46576.2022.9897219}}

@article{willard2023efficient,
  title   = {Efficient Guided Generation for LLMs},
  author  = {Willard, Brandon T and Louf, R{\'e}mi},
  journal = {arXiv preprint arXiv:2307.09702},
  year    = {2023}
}

@inproceedings{amadio2022target,
  title={Target-referred DMPS for learning bimanual tasks from shared-autonomy telemanipulation},
  author={Amadio, Fabio and Laghi, Marco and Raiano, Luigi and Rollo, Federico and Zunino, Andrea and Raiola, Gennaro and Ajoudani, Arash},
  booktitle={IEEE-RAS International Conference on Humanoid Robots (Humanoids)},
  pages={496--503},
  year={2022}
}

@inproceedings{labbe2022megapose,
title={MegaPose: 6D Pose Estimation of Novel Objects via Render \& Compare},
author={Yann Labb{\'e} and Lucas Manuelli and Arsalan Mousavian and Stephen Tyree and Stan Birchfield and Jonathan Tremblay and Justin Carpentier and Mathieu Aubry and Dieter Fox and Josef Sivic},
booktitle={Conference on Robot Learning (CoRL)},
year={2022},
}

@misc{grattafiori2024llama3herdmodels,
      title={The Llama 3 Herd of Models}, 
      author={Aaron Grattafiori and Abhimanyu Dubey and Abhinav Jauhri and Abhinav Pandey and Abhishek Kadian and Ahmad Al-Dahle and Aiesha Letman and Akhil Mathur and Alan Schelten and Alex Vaughan and Amy Yang and Angela Fan and Anirudh Goyal and Anthony Hartshorn and Aobo Yang and Archi Mitra and Archie Sravankumar and Artem Korenev and Arthur Hinsvark and Arun Rao and Aston Zhang and Aurelien Rodriguez and Austen Gregerson and Ava Spataru and Baptiste Roziere and Bethany Biron and Binh Tang and Bobbie Chern and Charlotte Caucheteux and Chaya Nayak and Chloe Bi and Chris Marra and Chris McConnell and Christian Keller and Christophe Touret and Chunyang Wu and Corinne Wong and Cristian Canton Ferrer and Cyrus Nikolaidis and Damien Allonsius and Daniel Song and Danielle Pintz and Danny Livshits and Danny Wyatt and David Esiobu and Dhruv Choudhary and Dhruv Mahajan and Diego Garcia-Olano and Diego Perino and Dieuwke Hupkes and Egor Lakomkin and Ehab AlBadawy and Elina Lobanova and Emily Dinan and Eric Michael Smith and Filip Radenovic and Francisco Guzmán and Frank Zhang and Gabriel Synnaeve and Gabrielle Lee and Georgia Lewis Anderson and Govind Thattai and Graeme Nail and Gregoire Mialon and Guan Pang and Guillem Cucurell and Hailey Nguyen and Hannah Korevaar and Hu Xu and Hugo Touvron and Iliyan Zarov and Imanol Arrieta Ibarra and Isabel Kloumann and Ishan Misra and Ivan Evtimov and Jack Zhang and Jade Copet and Jaewon Lee and Jan Geffert and Jana Vranes and Jason Park and Jay Mahadeokar and Jeet Shah and Jelmer van der Linde and Jennifer Billock and Jenny Hong and Jenya Lee and Jeremy Fu and Jianfeng Chi and Jianyu Huang and Jiawen Liu and Jie Wang and Jiecao Yu and Joanna Bitton and Joe Spisak and Jongsoo Park and Joseph Rocca and Joshua Johnstun and Joshua Saxe and Junteng Jia and Kalyan Vasuden Alwala and Karthik Prasad and Kartikeya Upasani and Kate Plawiak and Ke Li and Kenneth Heafield and Kevin Stone and Khalid El-Arini and Krithika Iyer and Kshitiz Malik and Kuenley Chiu and Kunal Bhalla and Kushal Lakhotia and Lauren Rantala-Yeary and Laurens van der Maaten and Lawrence Chen and Liang Tan and Liz Jenkins and Louis Martin and Lovish Madaan and Lubo Malo and Lukas Blecher and Lukas Landzaat and Luke de Oliveira and Madeline Muzzi and Mahesh Pasupuleti and Mannat Singh and Manohar Paluri and Marcin Kardas and Maria Tsimpoukelli and Mathew Oldham and Mathieu Rita and Maya Pavlova and Melanie Kambadur and Mike Lewis and Min Si and Mitesh Kumar Singh and Mona Hassan and Naman Goyal and Narjes Torabi and Nikolay Bashlykov and Nikolay Bogoychev and Niladri Chatterji and Ning Zhang and Olivier Duchenne and Onur Çelebi and Patrick Alrassy and Pengchuan Zhang and Pengwei Li and Petar Vasic and Peter Weng and Prajjwal Bhargava and Pratik Dubal and Praveen Krishnan and Punit Singh Koura and Puxin Xu and Qing He and Qingxiao Dong and Ragavan Srinivasan and Raj Ganapathy and Ramon Calderer and Ricardo Silveira Cabral and Robert Stojnic and Roberta Raileanu and Rohan Maheswari and Rohit Girdhar and Rohit Patel and Romain Sauvestre and Ronnie Polidoro and Roshan Sumbaly and Ross Taylor and Ruan Silva and Rui Hou and Rui Wang and Saghar Hosseini and Sahana Chennabasappa and Sanjay Singh and Sean Bell and Seohyun Sonia Kim and Sergey Edunov and Shaoliang Nie and Sharan Narang and Sharath Raparthy and Sheng Shen and Shengye Wan and Shruti Bhosale and Shun Zhang and Simon Vandenhende and Soumya Batra and Spencer Whitman and Sten Sootla and Stephane Collot and Suchin Gururangan and Sydney Borodinsky and Tamar Herman and Tara Fowler and Tarek Sheasha and Thomas Georgiou and Thomas Scialom and Tobias Speckbacher and Todor Mihaylov and Tong Xiao and Ujjwal Karn and Vedanuj Goswami and Vibhor Gupta and Vignesh Ramanathan and Viktor Kerkez and Vincent Gonguet and Virginie Do and Vish Vogeti and Vítor Albiero and Vladan Petrovic and Weiwei Chu and Wenhan Xiong and Wenyin Fu and Whitney Meers and Xavier Martinet and Xiaodong Wang and Xiaofang Wang and Xiaoqing Ellen Tan and Xide Xia and Xinfeng Xie and Xuchao Jia and Xuewei Wang and Yaelle Goldschlag and Yashesh Gaur and Yasmine Babaei and Yi Wen and Yiwen Song and Yuchen Zhang and Yue Li and Yuning Mao and Zacharie Delpierre Coudert and Zheng Yan and Zhengxing Chen and Zoe Papakipos and Aaditya Singh and Aayushi Srivastava and Abha Jain and Adam Kelsey and Adam Shajnfeld and Adithya Gangidi and Adolfo Victoria and Ahuva Goldstand and Ajay Menon and Ajay Sharma and Alex Boesenberg and Alexei Baevski and Allie Feinstein and Amanda Kallet and Amit Sangani and Amos Teo and Anam Yunus and Andrei Lupu and Andres Alvarado and Andrew Caples and Andrew Gu and Andrew Ho and Andrew Poulton and Andrew Ryan and Ankit Ramchandani and Annie Dong and Annie Franco and Anuj Goyal and Aparajita Saraf and Arkabandhu Chowdhury and Ashley Gabriel and Ashwin Bharambe and Assaf Eisenman and Azadeh Yazdan and Beau James and Ben Maurer and Benjamin Leonhardi and Bernie Huang and Beth Loyd and Beto De Paola and Bhargavi Paranjape and Bing Liu and Bo Wu and Boyu Ni and Braden Hancock and Bram Wasti and Brandon Spence and Brani Stojkovic and Brian Gamido and Britt Montalvo and Carl Parker and Carly Burton and Catalina Mejia and Ce Liu and Changhan Wang and Changkyu Kim and Chao Zhou and Chester Hu and Ching-Hsiang Chu and Chris Cai and Chris Tindal and Christoph Feichtenhofer and Cynthia Gao and Damon Civin and Dana Beaty and Daniel Kreymer and Daniel Li and David Adkins and David Xu and Davide Testuggine and Delia David and Devi Parikh and Diana Liskovich and Didem Foss and Dingkang Wang and Duc Le and Dustin Holland and Edward Dowling and Eissa Jamil and Elaine Montgomery and Eleonora Presani and Emily Hahn and Emily Wood and Eric-Tuan Le and Erik Brinkman and Esteban Arcaute and Evan Dunbar and Evan Smothers and Fei Sun and Felix Kreuk and Feng Tian and Filippos Kokkinos and Firat Ozgenel and Francesco Caggioni and Frank Kanayet and Frank Seide and Gabriela Medina Florez and Gabriella Schwarz and Gada Badeer and Georgia Swee and Gil Halpern and Grant Herman and Grigory Sizov and Guangyi and Zhang and Guna Lakshminarayanan and Hakan Inan and Hamid Shojanazeri and Han Zou and Hannah Wang and Hanwen Zha and Haroun Habeeb and Harrison Rudolph and Helen Suk and Henry Aspegren and Hunter Goldman and Hongyuan Zhan and Ibrahim Damlaj and Igor Molybog and Igor Tufanov and Ilias Leontiadis and Irina-Elena Veliche and Itai Gat and Jake Weissman and James Geboski and James Kohli and Janice Lam and Japhet Asher and Jean-Baptiste Gaya and Jeff Marcus and Jeff Tang and Jennifer Chan and Jenny Zhen and Jeremy Reizenstein and Jeremy Teboul and Jessica Zhong and Jian Jin and Jingyi Yang and Joe Cummings and Jon Carvill and Jon Shepard and Jonathan McPhie and Jonathan Torres and Josh Ginsburg and Junjie Wang and Kai Wu and Kam Hou U and Karan Saxena and Kartikay Khandelwal and Katayoun Zand and Kathy Matosich and Kaushik Veeraraghavan and Kelly Michelena and Keqian Li and Kiran Jagadeesh and Kun Huang and Kunal Chawla and Kyle Huang and Lailin Chen and Lakshya Garg and Lavender A and Leandro Silva and Lee Bell and Lei Zhang and Liangpeng Guo and Licheng Yu and Liron Moshkovich and Luca Wehrstedt and Madian Khabsa and Manav Avalani and Manish Bhatt and Martynas Mankus and Matan Hasson and Matthew Lennie and Matthias Reso and Maxim Groshev and Maxim Naumov and Maya Lathi and Meghan Keneally and Miao Liu and Michael L. Seltzer and Michal Valko and Michelle Restrepo and Mihir Patel and Mik Vyatskov and Mikayel Samvelyan and Mike Clark and Mike Macey and Mike Wang and Miquel Jubert Hermoso and Mo Metanat and Mohammad Rastegari and Munish Bansal and Nandhini Santhanam and Natascha Parks and Natasha White and Navyata Bawa and Nayan Singhal and Nick Egebo and Nicolas Usunier and Nikhil Mehta and Nikolay Pavlovich Laptev and Ning Dong and Norman Cheng and Oleg Chernoguz and Olivia Hart and Omkar Salpekar and Ozlem Kalinli and Parkin Kent and Parth Parekh and Paul Saab and Pavan Balaji and Pedro Rittner and Philip Bontrager and Pierre Roux and Piotr Dollar and Polina Zvyagina and Prashant Ratanchandani and Pritish Yuvraj and Qian Liang and Rachad Alao and Rachel Rodriguez and Rafi Ayub and Raghotham Murthy and Raghu Nayani and Rahul Mitra and Rangaprabhu Parthasarathy and Raymond Li and Rebekkah Hogan and Robin Battey and Rocky Wang and Russ Howes and Ruty Rinott and Sachin Mehta and Sachin Siby and Sai Jayesh Bondu and Samyak Datta and Sara Chugh and Sara Hunt and Sargun Dhillon and Sasha Sidorov and Satadru Pan and Saurabh Mahajan and Saurabh Verma and Seiji Yamamoto and Sharadh Ramaswamy and Shaun Lindsay and Shaun Lindsay and Sheng Feng and Shenghao Lin and Shengxin Cindy Zha and Shishir Patil and Shiva Shankar and Shuqiang Zhang and Shuqiang Zhang and Sinong Wang and Sneha Agarwal and Soji Sajuyigbe and Soumith Chintala and Stephanie Max and Stephen Chen and Steve Kehoe and Steve Satterfield and Sudarshan Govindaprasad and Sumit Gupta and Summer Deng and Sungmin Cho and Sunny Virk and Suraj Subramanian and Sy Choudhury and Sydney Goldman and Tal Remez and Tamar Glaser and Tamara Best and Thilo Koehler and Thomas Robinson and Tianhe Li and Tianjun Zhang and Tim Matthews and Timothy Chou and Tzook Shaked and Varun Vontimitta and Victoria Ajayi and Victoria Montanez and Vijai Mohan and Vinay Satish Kumar and Vishal Mangla and Vlad Ionescu and Vlad Poenaru and Vlad Tiberiu Mihailescu and Vladimir Ivanov and Wei Li and Wenchen Wang and Wenwen Jiang and Wes Bouaziz and Will Constable and Xiaocheng Tang and Xiaojian Wu and Xiaolan Wang and Xilun Wu and Xinbo Gao and Yaniv Kleinman and Yanjun Chen and Ye Hu and Ye Jia and Ye Qi and Yenda Li and Yilin Zhang and Ying Zhang and Yossi Adi and Youngjin Nam and Yu and Wang and Yu Zhao and Yuchen Hao and Yundi Qian and Yunlu Li and Yuzi He and Zach Rait and Zachary DeVito and Zef Rosnbrick and Zhaoduo Wen and Zhenyu Yang and Zhiwei Zhao and Zhiyu Ma},
      year={2024},
      eprint={2407.21783},
      archivePrefix={arXiv},
      primaryClass={cs.AI},
      url={https://arxiv.org/abs/2407.21783}, 
}

@inproceedings{rouxel2024flow,
  title={Flow matching imitation learning for multi-support manipulation},
  author={Rouxel, Quentin and Ferrari, Andrea and Ivaldi, Serena and Mouret, Jean-Baptiste},
  booktitle={IEEE-RAS International Conference on Humanoid Robots (Humanoids)},
  pages={528--535},
  year={2024}
}

@article{coucke2018snips,
  title   = {Snips Voice Platform: an embedded Spoken Language Understanding system for private-by-design voice interfaces},
  author  = {Coucke, Alice and Saade, Alaa and Ball, Adrien and Bluche, Th{\'e}odore and Caulier, Alexandre and Leroy, David and Doumouro, Cl{\'e}ment and Gisselbrecht, Thibault and Caltagirone, Francesco and Lavril, Thibaut and others},
  journal = {arXiv preprint arXiv:1805.10190},
  pages   = {12--16},
  year    = {2018}
}

@misc{Eren_Coqui_TTS_2021,
author = {Eren, Gölge and {The Coqui TTS Team}},
doi = {10.5281/zenodo.6334862},
license = {MPL-2.0},
month = jan,
title = {{Coqui TTS}},
url = {https://github.com/coqui-ai/TTS},
version = {1.4},
year = {2021}
}

@article{chi2023diffusion,
  title={Diffusion policy: Visuomotor policy learning via action diffusion},
  author={Chi, Cheng and Xu, Zhenjia and Feng, Siyuan and Cousineau, Eric and Du, Yilun and Burchfiel, Benjamin and Tedrake, Russ and Song, Shuran},
  journal={The International Journal of Robotics Research},
  year={2023},
  publisher={SAGE Publications Sage UK: London, England}
}

@inproceedings{wei2023chainofthought,
    title={Chain of Thought Prompting Elicits Reasoning in Large Language Models},
    author={Jason Wei and Xuezhi Wang and Dale Schuurmans and Maarten Bosma and Brian Ichter and Fei Xia and Ed H. Chi and Quoc V Le and Denny Zhou},
    booktitle={Advances in Neural Information Processing Systems},
    year={2022}
}

@inproceedings{brown2020language,
  title         = {Language Models are Few-Shot Learners},
  author        = {Tom B. Brown and Benjamin Mann and Nick Ryder and Melanie Subbiah and Jared Kaplan and Prafulla Dhariwal and Arvind Neelakantan and Pranav Shyam and Girish Sastry and Amanda Askell and Sandhini Agarwal and Ariel Herbert-Voss and Gretchen Krueger and Tom Henighan and Rewon Child and Aditya Ramesh and Daniel M. Ziegler and Jeffrey Wu and Clemens Winter and Christopher Hesse and Mark Chen and Eric Sigler and Mateusz Litwin and Scott Gray and Benjamin Chess and Jack Clark and Christopher Berner and Sam McCandlish and Alec Radford and Ilya Sutskever and Dario Amodei},
  year          = {2020},
  booktitle     = {Advances in Neural Information Processing Systems}
}

@inproceedings{wang2016apriltag,
  title={AprilTag 2: Efficient and robust fiducial detection},
  author={Wang, John and Olson, Edwin},
  booktitle={IEEE/RSJ International Conference on Intelligent Robots and Systems (IROS)},
  year={2016},
}

@inproceedings{krogius2019flexible,
  title={Flexible layouts for fiducial tags},
  author={Krogius, M. and Haggenmiller, A. and Olson, E.},
  booktitle={IEEE/RSJ International Conference on Intelligent Robots and Systems (IROS)},
  year={2019},
}

@ARTICLE{9131861,
  author={Li, Gaofeng and Caponetto, Fernando and Del Bianco, Edoardo and Katsageorgiou, Vasiliki and Sarakoglou, Ioannis and Tsagarakis, Nikos G.},
  journal={IEEE Robotics and Automation Letters}, 
  title={Incomplete Orientation Mapping for Teleoperation With One DoF Master-Slave Asymmetry}, 
  year={2020},
  volume={5},
  number={4},
  pages={5167-5174},
  doi={10.1109/LRA.2020.3006796}}

@INPROCEEDINGS{8700380,
  author={Carpentier, Justin and Saurel, Guilhem and Buondonno, Gabriele and Mirabel, Joseph and Lamiraux, Florent and Stasse, Olivier and Mansard, Nicolas},
  booktitle={2019 IEEE/SICE International Symposium on System Integration (SII)}, 
  title={The Pinocchio C++ library : A fast and flexible implementation of rigid body dynamics algorithms and their analytical derivatives}, 
  year={2019},
  volume={},
  number={},
  pages={614-619},
  doi={10.1109/SII.2019.8700380}}

@inproceedings{totsila2025safe,
  TITLE = {{Safe Bimanual Teleoperation with Language-Guided Collision Avoidance}},
  AUTHOR = {Totsila, Dionis and Donoso, Clemente and Hoffman, Enrico Mingo and Mouret, Jean-Baptiste and Ivaldi, Serena},
  URL = {https://inria.hal.science/hal-05123517},
  BOOKTITLE = {{2025 IEEE Conference on Telepresence}},
  ADDRESS = {Leiden, Netherlands},
  YEAR = {2025},
  MONTH = Sep,
  HAL_ID = {hal-05123517},
  HAL_VERSION = {v1},
}

\end{document}